# COMPARING EXPERT SYSTEMS
# BUILT USING DIFFERENT UNCERTAIN INFERENCE SYSTEMS[*]

David S. Vaughan
Bruce M. Perrin
Robert M. Yadrick

McDonnell Douglas Missile Systems Company
P.O. Box 516
St. Louis, MO 63166

Peter D. Holden

McDonnell Aircraft Company
P.O. Box 516
St. Louis, MO 63166

## ABSTRACT

This study compares the inherent intuitiveness or usability of the most prominent methods for managing uncertainty in expert systems, including those of EMYCIN, PROSPECTOR, Dempster-Shafer theory, Fuzzy set theory, simplified probability theory (assuming marginal independence), and linear regression using probability estimates. Participants in the study gained experience in a simple, hypothetical problem domain through a series of learning trials. They were then randomly assigned to develop an expert system using one of the six Uncertain Inference Systems (UISs) listed above. Performance of the resulting systems was then compared. The results indicate that the systems based on the PROSPECTOR and EMYCIN models were significantly less accurate for certain types of problems compared to systems based on the other UISs. Possible reasons for these differences are discussed.

## 1. INTRODUCTION

Several methods of managing uncertainty in expert systems, or Uncertain Inference Systems (UISs), have been proposed or developed over the past twenty years, including those devised for the MYCIN [1] and PROSPECTOR [2] projects, methods based on fuzzy set theory [3] and Dempster-Shafer theory [4], approaches using simplifications of probability theory [e.g., assuming marginal independence; 5], among others. However, until recent years, little information, apart from theoretical or philosophical arguments, has been available to aid in the selection of a UIS for a particular application. Empirical studies comparing the algorithmic accuracy of UISs began to appear about four years ago [e.g., 6, 7, 8], and later methodical improvements [9, 10] resulted in identification of the relative limitations of several of these UISs under optimal conditions.

Many of these approximate techniques were justified by their developers, at least in part, however, by arguing that their approach was more compatible with human mental representations and/or reasoning, more intuitive, and thus, more practical to use. Heuristic parameters and ad hoc combining methods which result in algorithmic inaccuracy might be offset, the argument goes, if humans can readily and accurately apply the approach to a given problem. These claims of enhanced "usability" of the various UIS are virtually untested.

Henrion and Cooley [11] and Heckerman [12] have reported single-expert case studies that compared UISs which were used to develop relatively large, complex applications. Such case studies provide useful insights into the system development process; however, it is difficult to separate the effect of the usability of the approach from

---

the numerous uncontrolled factors in an actual application. Some of these uncontrolled factors include 1) variation in factors other than the choice of a UIS, such as knowledge acquisition techniques, user interface, development environments, and the like, which are not an inherent feature of the UIS; 2) interference produced by requiring a single expert to provide multiple forms of parameter estimates and expectations or preferences of the expert for one of these forms over another; and 3) differences in the expectations, or where multiple individuals were involved, abilities of the system developers. Additionally, it is difficult to know how to generalize from the experience of a single individual to the broader issue of usability of a UIS or what can serve as an criterion of accuracy independent of the judgments of the application expert. In all fairness to the authors of these studies, a comparison of the inherent usability of the UIS was not the primary focus of their work.

Only a single study, one by Mitchell, Harp, and Simkin [13], was found which compared UIS under controlled learning and acquisition conditions. Unfortunately, participants in their study did not actually develop a system, but rather, estimated parameters which might be used in a system, such as a EMYCIN Certainty Factor (CF) for a change in belief in a conclusion given evidence. Thus, their participants were given no opportunity to observe the results of their parameter estimates, in terms of a system answer, and refine the estimates. Additionally, their study did not involve uncertainty in the observations of evidence, e.g., assigning a CF to the observed evidence, which makes drawing implications for uncertainty management even more problematic.

The current study sought to control factors extraneous to inherent UIS differences as fully as practical, while providing all of the features necessary to build, test, and refine a working expert system. Differences in expertise were controlled by training the participants in a hypothetical problem domain to a criterion level of performance. Instructions to the participants were standardized and every effort was made to fully describe the essential features of the UIS without giving information that might influence individual parameter values or rules. The entire procedure, apart from the instructions and a questionnaire, was automated; the system development environments for each of the UISs provided only simple editors and displays to minimize differences in knowledge acquisition and interface technologies. In short, the present study addressed the question of inherent usability of a UIS by examining how readily and accurately individuals could use the essential features of a given UIS.

## 2. METHOD

Participants in this study were 60 volunteers from McDonnell Douglas Corporation interested or involved in AI activities. They varied widely in their knowledge of AI and in their experience with expert systems, ranging from relatively experienced knowledge engineers to managers of AI projects to personnel just introduced to the technology. Ten of these volunteers were randomly assigned to work with each of the six different UISs examined in this study.

2.1 Learning Trials

To assure comparable levels of expertise, the participants in the study gained experience in solving a hypothetical diagnostic problem through a series of learning trials. Specifically, the problem was to diagnose whether a fictitious machine was or was not malfunctioning according to temperature and pressure readings. During the learning trials, participants saw a temperature and a pressure on a computer terminal and responded by typing an "M" for malfunction or a "W" for working, depending on which they believed most likely given the readings. They were then informed of the correct answer for that particular case, and a new temperature/pressure was presented to begin a new trial. This process continued until the participant achieved an average level of accuracy over 20 consecutive trials equal to about 85% correct, near optimal performance given the inherent unpredictability of the outcome.

Problems for the learning (and test trials, which are described later) were generated in accordance with the probability of their occurrence, as indicated by the contingency table given in Table 1. For example, problems for which both temperature and pressure were high (abnormal) and the machine was malfunctioning were generated, on average, in 0.315 of the cases. The conditional probabilities of the conclusion (malfunction) given the various states of the evidence are as follows: $P(C|\sim E1\&\sim E2)$ = probability of malfunction given normal temperature and normal pressure = 0.1; $P(C|\sim E1\&E2) = P(C|E1\&\sim E2) = 0.2$; and $P(C|E1\&E2) = 0.9$. The table is therefore conjunctive in nature, in that a malfunction is likely when both temperature and pressure are high (0.9) and unlikely otherwise (0.2 or less).

| TABLE 1. THE CONTINGENCY TABLE FROM WHICH PROBLEMS WERE GENERATED ||
|---|---|
| EVIDENCE | CONCLUSION |
|  | WORKING (~C)    MALFUNCTION (C) |
| NORMAL TEMPERATURE (~E1)/ NORMAL PRESSURE (~E2) | 0.315          0.035 |
| NORMAL TEMPERATURE (~E1)/ HIGH PRESSURE (E2) | 0.120          0.030 |
| HIGH TEMPERATURE (E1)/ NORMAL PRESSURE (~E2) | 0.120          0.030 |
| HIGH TEMPERATURE (E1)/ HIGH PRESSURE (E2) | 0.035          0.315 |

The actual temperature and pressure readings were generated after it had been determined which cell of the contingency table a particular problem was to represent. We sampled from normal distributions with different means depending on whether the problem represented a normal or a high reading. Means and standard deviations for the normal and high temperature and pressure distributions are given in Table 2. Thus, although problem presentation was controlled by a contingency table and sampling from normal distributions, there was nothing in the display or the instructions that suggested one interpretation of the problem was more appropriate than another. Problem displays and the instructions were intentionally vague on issues such as whether the abnormal temperature and pressure readings were symptoms or causes of the malfunction, whether or not the uncertainty was primarily in the readings (e.g., sensor unreliability) or in the relation between the readings and the conclusion, and what the exact likelihood of malfunction was when neither, either, or both readings were abnormal.

| TABLE 2. MEANS AND STANDARD DEVIATIONS FOR NORMAL AND HIGH DISTRIBUTIONS ||
|---|---|
|  | MEAN    STD. DEV. |
| NORMAL TEMPERATURE | 180.0      5.0 |
| HIGH TEMPERATURE | 200.0      5.0 |
| NORMAL PRESSURE | 70.0      3.0 |
| HIGH PRESSURE | 82.0      3.0 |

2.2 System Building & Tuning Trials

Study participants then used one of six UISs to build, test, and refine a system that captured their knowledge about the diagnosis problem. The study included UIS implementations based on the EMYCIN and PROSPECTOR models, simplified probability theory (assuming marginal independence of the evidence), linear regression using probabilities, fuzzy set theory, and Dempster-Shafer theory. The UIS's relevant features were described and illustrated in written instructions, and every attempt was made to make the descriptions for the different UISs comparable in the amount and level of detail of the information. The following overview of the approaches we studied is only intended to provide an appreciation of some of the major differences between the UISs; for a full explication of the models, the interested reader should consult the references cited.

    (a) EMYCIN & PROSPECTOR. Although different in computation and parameters, the EMYCIN [1] and PROSPECTOR [2] UISs have similar rule formats. Both UISs allow simple rules, which relate one piece of evidence with the conclusion, and conjunctive ("AND") and disjunctive ("OR") rules, which relate a particular type

of evidence combination to the conclusion. Each of these UISs requires estimation of one or more parameters for each rule (EMYCIN requires a single parameter, a CF to specify the strength of the link between evidence and conclusion, while PROSPECTOR requires two, a Logical Sufficiency and a Logical Necessity). CFs are estimated on a -1 to +1 scale, while the PROSPECTOR parameters are estimated on a +6 to -6 scale. Additionally, the participants using the PROSPECTOR UIS must estimate a prior odds for the evidence of each rule and the conclusion, this on a scale of +4 to -4. Verbal descriptors of these scale values have been provided by the developers and were available to the study participants.

(b) Linear Regression & Simplified Probability (Independence). Equations (1) and (2) below define the regression and simplified probability (or independence) models, respectively. The parameters for both are probabilities, expressed on a zero-to-one scale (the procedure implemented actually requested proportions to express relations).

$$P'(C) = a + b1 * P'(E1) + b2 * P'(E2) \tag{1}$$

$$\begin{aligned}P'(C) = &P'(\sim E1) * P'(\sim E2) * P(C|\sim E1 \& \sim E2) + \\ &P'(\sim E1) * P'(E2) * P(C|\sim E1 \& E2) + \\ &P'(E1) * P'(\sim E2) * P(C|E1 \& \sim E2) + \\ &P'(E1) * P'(E2) * P(C|E1 \& E2)\end{aligned} \tag{2}$$

The regression UIS required an estimated proportion of cases that would have a malfunction when neither piece of evidence was present, corresponding to the intercept "a" in equation (1), and the proportion when each piece of evidence was present, corresponding to the weights "b1" and "b2". The independence UIS required an estimate of the proportions when neither, both, and either piece of evidence alone was present, for a total of four parameters. Nothing corresponding to rule selection is required under these approaches, as the form of the relations is specified by the models.

(c) Fuzzy Set & Dempster-Shafer Theory. Both of these UISs represent broad conceptual theories of uncertainty, and so, they present a variety of alternative implementations. In both cases, we chose one of the more simple, straightforward implementations, but one we believed was adequate for the problem. Certainly, additional empirical research into these alternatives is warranted.

Our implementation of fuzzy set theory [3] used fuzzy membership functions and rule specification. Participants using this UIS described a membership function which mapped particular evidence values onto their rules. The membership function required eight parameters for each type of evidence. Four of these parameters indicated the high and low values of an interval over which the evidence was deemed definitely present (the bounded values of temperature or pressure were definitely "high") and the high and low points of an interval over which the evidence was deemed definitely absent. The remaining four parameters indicated the interval over which the evidence was considered uncertain. Rules were either simple or conjunctive, and the strength of the rules was expressed as a probability. These forms of membership functions and rule sets are based loosely on those described and illustrated by Bonissone and Decker [14].

Our implementation of Dempster-Shafer theory [4] used simple support functions combined using Dempster's rule. The frame of discernment for the problem in this study was the set {working, malfunctioning} and each reading (temperature or pressure) was taken to be compatible with one, but not both of these conclusions.. For example, a reading that supported the conclusion that the mechanism was malfunctioning did not support the conclusion that it was working. To estimate the support for a conclusion given a reading, participants were asked to estimate the readings which supported the following beliefs:
        Bel(working) = 0.999, Bel(malfunction) = 0.0;
        Bel(working) = 0.50, Bel(malfunction) = 0.0;
        Bel(working) = 0.0, Bel(malfunction) = 0.0;
        Bel(working) = 0.0, Bel(malfunction) = 0.50; and
        Bel(working) = 0.0, Bel(malfunction) = 0.999.
These points defined a support function, and linear interpolation was used to estimate beliefs for evidence readings falling between those points supplied by the participants. Finally, Dempster's rule was used to combine beliefs.

Once the systems were developed, the participants tested them using temperature and pressure values of their own choosing. If the system answer, which was in a form appropriate to the UIS, disagreed with their own evaluation, simple editors were available to modify any of the UIS's features. Testing and refining continued iteratively until the participant was satisfied with the system's performance.

2.3 Test Trials

Finally, each system was used to diagnose thirty test cases and the participants completed a brief questionnaire describing their background and their impressions of the UIS they had used. Throughout the study, participants were encouraged to ask any questions they wished. Questions requesting clarification of procedures and the like were answered at any time to the participant's satisfaction. If a participant asked a question pertaining specifically to the problem domain or to implementing their solution in a given UIS, the experimenter explained that figuring such things out was part of the task at hand.

3. RESULTS AND DISCUSSION

Tests of mean differences suggest that the experimental procedure was effective in creating groups with roughly equivalent backgrounds and understanding of the problem domain. Analysis of variance (ANOVA) tests of the background characteristics, which included items on exposure to AI and expert systems methodologies, revealed no significant differences between the groups assigned to use the different UISs [maximum $F(5,53) = 0.88$, $p < 0.50$]. Likewise, we found no differences between groups in the number of trials taken to reach the learning criterion [$F(5,54) = 0.21$, $p < 0.96$].

We also tested for the clarity or ease of use of the UISs in two ways: 1) in the self-reported ratings of comprehensiveness, consistency, and ease of use of the UIS from the questionnaire administered at the end of the session; and 2) in the effort expended to tune a given UIS to the satisfaction of the participant. ANOVA tests indicated the groups were not significantly different in the self-reported ratings [maximum $F(5,53) = 1.56$, $p < 0.19$]. This finding may imply that the UISs are equivalent in terms of clarity. It is also possible, however, that the participants, who used only one of the UIS during the study, lacked any basis for comparison.

Differences in the average number of trials the participants spent in tuning the different UISs are given in Table 3. An ANOVA indicated that the groups differed significantly [$F(5,54) = 2.63$, $p < 0.04$], with participants using the linear regression approach requiring the fewest trials, while those applying the PROSPECTOR UIS needing more than twice as many trials, on the average. The difference in the number of tuning trials may reflect the relative clarity or ease of use of the approaches; however, it may simply be related to the number of distinct parameters that must be estimated. The linear regression approach involves the estimation of the fewest parameters, only three, while participants using PROSPECTOR estimated an average of 13.0 parameters. Once this variance (in the number of parameters) was taken into account using analysis of covariance, the groups did not differ significantly in the number of trials to tune their systems [$F(5,53) = 0.42$, $p < 0.83$].

| TABLE 3. NUMBER OF TRIALS TO TUNE A GIVEN UNCERTAIN INFERENCE SYSTEM | | | | | | |
|---|---|---|---|---|---|---|
| | | | GROUP | | | |
| | EMYCIN | PROSP. | INDEP. | LNR REG. | FUZZY SET | D-SHAFER |
| NO. OF TRIALS | 9.5 | 19.3 | 11.3 | 8.1 | 18.5 | 10.0 |

We evaluated six different measures to assess the accuracy of the participants' systems on the final thirty test trials; however, all six revealed the same pattern of findings. Consequently, for simplicity, we will report results based on the proportion of correct diagnoses by the system compared to the answer used to generate the problem (i.e., the cell of the contingency table sampled). The ANOVA for this index showed a significant main effect for trials and a significant trials by UIS interaction. The summary ANOVA table for this analysis is given as Table 4.

The significant trials by UIS interaction [$F(145,1566) = 1.58$, $p < 0.01$] indicates substantial variation among the UISs in their accuracy given different types of trials.

TABLE 4. SUMMARY ANOVA TABLE FOR PROPORTION CORRECT DIAGNOSES

| SOURCE | DF | SS | MS | F |
|---|---|---|---|---|
| BETWEEN | 59 | 52.53 | | |
| UIS | 5 | 5.41 | 1.08 | 1.24 |
| SUBJ(UIS) | 54 | 47.12 | 0.87 | |
| WITHIN | 1740 | 269.47 | | |
| TRIALS | 29 | 71.50 | 2.47 | 22.36* |
| UIS*TRIALS | 145 | 25.29 | 0.17 | 1.58* |
| TRIALS*S(UIS) | 1566 | 172.68 | 0.11 | |
| TOTAL | 1799 | 322.0 | | |

*$P < 0.01$

Further evaluation of the trial by UIS interaction revealed that, on average, systems developed using the six UISs were equally accurate given "consistent" evidence, that is, when both temperature and pressure readings were normal or both were high. However, performance of the different UISs varied widely when the evidence was "mixed" or conflicting, i.e., one of the readings was high while the other was normal. These results are summarized in Table 5. A Tukey test of these mean differences indicated that the systems developed using the EMYCIN and PROSPECTOR UISs were significantly less accurate given mixed evidence than they, or any of the UISs were, when given consistent evidence.

TABLE 5. PROPORTION CORRECT DIAGNOSES GIVEN DIFFERENT TYPES OF EVIDENCE

| | EMYCIN | PROSP. | GROUP INDEP. | LNR REG. | FUZZY SET | D-SHAFER |
|---|---|---|---|---|---|---|
| CONSISTENT EVID. | 0.90 | 0.85 | 0.87 | 0.85 | 0.84 | 0.93 |
| MIXED EVIDENCE | 0.31 | 0.26 | 0.74 | 0.53 | 0.57 | 0.60 |

There are several possible explanations why systems built using the EMYCIN and PROSPECTOR UISs might be less accurate given mixed evidence. Although it is not likely, perhaps the developers of these systems had an incorrect impression of the likelihood of a malfunction given mixed information -- an impression not held by developers of the other types of systems. A second possibility is that system developers using the EMYCIN or PROSPECTOR UISs failed to write rules covering the mixed evidence cases or the rules they wrote were internally inconsistent. Third, although the rule base may be complete and consistent, it might be the case that the parameter values expressing the strength of the relations were inaccurately estimated. Finally, given equally well developed rule bases and equally accurate parameter estimates for the problem domain, it may be that the EMYCIN and PROSPECTOR UISs are less robust to the inaccuracies that exist in all the models.

We found support only for the third possibility, that the EMYCIN and PROSPECTOR parameters were inaccurately estimated. Table 6 shows the average parameter estimates (standardized on a zero to one scale) for each of the two types of mixed evidence cases -- normal temperature/high pressure or high temperature/normal pressure -- for each of the UISs. ANOVAs indicated significant differences among the UISs for both the normal temperature/high pressure parameter estimates [$F(5,54) = 8.75$, $p < 0.01$] and for the high pressure/normal pressure estimates [$F(5,54) = 5.20$, $p < 0.01$]. In general, the mixed case parameter values for EMYCIN and PROSPECTOR were inflated compared to the other UISs. Additionally, the PROSPECTOR and EMYCIN based systems were found to overestimate the likelihood of malfunction in mixed cases, even though the participants using these UISs were found to be no more likely than other participants to guess malfunction given mixed evidence in the learning trials.

| TABLE 6. AVERAGE UIS PARAMETER ESTIMATES FOR MIXED EVIDENCE CASES | | | | | | |
|---|---|---|---|---|---|---|
| | EMYCIN | PROSP. | GROUP INDEP. | LIN'R REG. | FUZZY SET | D-SHAFER |
| NORMAL TEMP/ HIGH PRESSURE | 0.81 | 0.80 | 0.43 | 0.33 | 0.49 | 0.50 |
| HIGH TEMP/ NORMAL PRESSURE | 0.59 | 0.71 | 0.40 | 0.37 | 0.35 | 0.50 |

## 4. CONCLUSIONS

Taken together, we believe the results of this research suggest that the rule-based approaches of EMYCIN and PROSPECTOR pose knowledge representation difficulties that are largely circumvented in the other, more structured methods. But before reviewing these implications, it is important to recognize some of the limitations of the research that is reported here. First, it may be that the knowledge captured in the present study is different in important ways from knowledge accrued through years of experience, the commonly recognized goal of an expert system. While it is not clear that a rule-based approach is a more appropriate or accurate method of representing that compiled knowledge, additional empirical research addressing the usability of the different UISs for encoding practical expertise is sorely needed.

A second possible limitation of the current study stems from the simple nature of the problem domain -- only two pieces of evidence were relevant to diagnosing a malfunction in the hypothetical mechanism. It is possible that in complex situations, where numerous sources of information bear on several different conclusions, that humans are capable of encoding only simple relations of the sort represented by the idealized conjunctive, disjunctive, or simple rules. Under this circumstance, the disadvantage of the rule-based UISs might disappear. Further research is also needed to evaluate this possibility.

Finally, the current study should not be interpreted as an indication of the potential or optimal performance of the different UISs. Clearly, all of the UISs evaluated in this study could be used to develop an accurate model of the simple problem domain; at least one very accurate system was developed using each of the different UISs in our study. Thus, this research does not necessarily reflect the final performance an experienced system developer might be able to achieve using any of the approaches studied. The study does, in our opinion, reflect the intuitiveness or inherent usability of the various approaches and, to the extent that this influences important system building processes such as the transfer of information, the accuracy of the final application may be substantially affected.

The primary question raised by this research is why, when their performance in the learning trials indicated they knew the correct response, did participants using the EMYCIN and PROSPECTOR UISs build systems which overestimated the likelihood of malfunction in mixed evidence cases? While the present study has no data that directly bear on this question, the pattern of findings that emerged suggest that the developers using the EMYCIN and PROSPECTOR UISs found it difficult to tune simultaneously to both the typical, consistent evidence case and the atypical, mixed case. The evidence clearly suggests that the inaccuracy of the systems given mixed evidence was not due to oversight on the part of the developers. In fact, the EMYCIN and PROSPECTOR systems that contained explicit rules and parameter estimates for mixed evidence cases were, in general, less accurate than those that relied on default values for their estimate. Rather, it appears that the EMYCIN and PROSPECTOR developers found it difficult to identify and understand the interaction of parameters in their rule-based model of the problem domain.

## REFERENCES

1. Shortliffe, E.H. and Buchanan, B.G., A model of inexact reasoning in medicine, Mathematical Biosciences, 23, 1975, 351-379.